\documentclass[11pt]{article}

\usepackage{emnlp2021}
\usepackage{times}
\usepackage{latexsym}

\usepackage[utf8]{inputenc}

\usepackage{microtype}

\usepackage[noend]{algpseudocode}
\usepackage{algorithmicx,algorithm}
\usepackage{graphicx}
\usepackage{booktabs}
\usepackage{amssymb}
\usepackage{array}
\newcolumntype{P}[1]{>{\centering\arraybackslash}m{#1}}
\usepackage{makecell}
\usepackage{multirow}
\usepackage{hhline}
\usepackage{threeparttable}
\usepackage{mathtools}
\usepackage{amsmath}
\usepackage[symbol]{footmisc}

\title{\texttt{Lingxi}: A Diversity-aware Chinese Modern Poetry Generation System}

\author{Xinran Zhang\textsuperscript{1}, Maosong Sun\textsuperscript{12*}, Jiafeng Liu\textsuperscript{1}, Xiaobing Li\textsuperscript{1}\\
  \textsuperscript{1}Department of Music Artificial Intelligence and Music Information Technology \\Central Conservatory of Music, Beijing, China \\
  \textsuperscript{2}Department of Computer Science and Technology, Tsinghua University, Beijing, China\\
  Institute for Artificial Intelligence, Tsinghua University, Beijing, China\\
  State Key Lab on Intelligent Technology and Systems, Tsinghua University, Beijing, China\\
  \texttt{zhangxr.wspn@gmail.com, sms@tsinghua.edu.cn}
  }
\date{}
\begin{document}
\maketitle
\begin{abstract}
Poetry generation has been a difficult task in natural language processing. Unlike plain neural text generation tasks, poetry has a high requirement for novelty, since an easily-understood sentence with too many high frequency words might not be considered as poetic, while adequately ambiguous sentences with low frequency words can possibly be novel and creative. Inspired by this, we present \texttt{Lingxi}, a diversity-aware Chinese modern poetry generation system. We propose nucleus sampling with randomized head (NS-RH) algorithm, which randomizes the high frequency part (``head'') of the predicted distribution, in order to emphasize on the ``comparatively low frequency'' words. The proposed algorithm can significantly increase the novelty of generated poetry compared with traditional sampling methods. The permutation of distribution is controllable by tuning the filtering parameter that determines the ``head'' to permutate, achieving diversity-aware sampling. We find that even when a large portion of filtered vocabulary is randomized, it can actually generate fluent poetry but with notably higher novelty. We also propose a semantic-similarity-based rejection sampling algorithm, which creates longer and more informative context on the basis of the short input poetry title while maintaining high semantic similarity to the title, alleviating the off-topic issue.
\end{abstract}
\footnotetext[1]{Corresponding author.}

\section{Introduction}

Poetry generation has been a classical task in natural language processing (NLP) for many years. Recently, the auto-regressive Transformer model ~\citep{NIPS2017_7181} like GPT-2 ~\citep{radford2019language} as well as the pre-training/fine-tuning paradigm has become a benchmark for this task.

The primary challenge for this task is the high novelty requirement of poetry that differs from plain neural text generation task. As is widely applied in text generation, the nucleus sampling algorithm proposed by ~\citet{holtzman2019curious} uses stochastic sampling instead of beam search algorithm ~\citep{li-etal-2016-deep,NIPS2017_7259,wiseman-etal-2017-challenges} to alleviating the \emph{text degeneration} issue, meanwhile truncating the ``tail'' (low frequency part) of the distribution to guarantee quality. However, in poetry generation cases we find that these methods \emph{still} generate boring and even repetitive poetry, since they do not fully address the issue that human text does \emph{not} always favor high probability words ~\citep{holtzman2019curious}. This is of specific importance for poetry generation, since an easily-understood poetry (low perplexity with too many high frequency words) can be regarded as boring and of low quality, while poetry that contains an adequate amount of surprising and low frequency words can actually be poetic and creative.

Furthermore, in the cases of plain text generation tasks such as examples by ~\citet{holtzman2019curious}, the input context is comparatively long with detailed human-provided information, while predicting steps are fairly short (only 200 tokens, compared with the 1,024 maximum length of context in GPT-2). But in Chinese modern poetry generation scenario, we consider the case where the input context is only the poetry title or the keyword (short with less detailed information), while asking the model to predict a full passage of poetry. Empirically, this is a much more difficult task, and is more easily affected by text degeneration or resulting in tangency from the input title.

To address these challenges, we present \texttt{Lingxi}, a diversity-aware Chinese modern poetry generation system. Our contributions are as follows.

\begin{itemize}
  %\item For the CWS issue, we propose a segmentation algorithm that incorporates CWS in the tokenization process. This algorithm not only fully exploits the human knowledge in CWS, but also combines the advantage of frequency-based methods. It is able to achieve high coverage of the vocabulary while maintaining a reasonable vocabulary size.
  \item We propose \emph{nucleus sampling with randomized head} (NS-RH) algorithm. This algorithm randomizes the ``head'' (high frequency part) of the predicted distribution, and can increase the novelty of generated poetry. Surprisingly, we find that even if almost top $80\%$ of the vocabulary is randomized, it still generates fluent poetry but with notably higher novelty.
  \item We propose a \emph{semantic-similarity-based rejection sampling algorithm}. This algorithm can create longer and more informative context on the basis of the short input poetry title while maintaining high semantic similarity to the title, alleviating the off-topic issue.
\end{itemize}

%\section{System Design}

\section{Pre-trained Language Model}\label{preprocessing}

We release a pre-trained language model called \emph{GPT-2 LyricCN}. Following the pre-training/fine-tuning paradigm, we collect about 3,500 published books of Chinese novels as the pre-training corpus, aiming at novel-style literary language modeling. After that, we collect about 220,000 passages of Chinese modern poetry and lyrics as the fine-tuning corpus, aiming at transferring the knowledge of pre-trained language model to the specific domain of poetry composition. Both training phases utilize the auto-regressive Transformer with language model loss like GPT-2 ~\citep{radford2019language}.

%\subsection{``Long Tail'' Issue of the CWS Vocabulary}
For the Chinese vocabulary issue, most researches directly apply frequency-based methods such as byte pair encoding (BPE, ~\citealp{10.5555/177910.177914,sennrich-etal-2016-neural}) or unigram language model (Unigram LM, ~\citealp{kudo-2018-subword}). However, Chinese word segmentation (CWS) is naturally a classical Chinese NLP task, which can be modeled by supervised learning ~\citep{
liu-etal-2014-domain,yan-etal-2020-graph,qiu-etal-2020-concise,duan-zhao-2020-attention}. Inspired by this, we propose a heuristic method that incorporates CWS in the preprocessing of the corpus. We first adopt a CWS tool known as \texttt{THULAC} by ~\citet{sun2016thulac} to process the corpus. This will generate a very large vocabulary that consists of unique segmented words with a very long low frequency ``tail''. To handle this, we propose the following heuristic algorithm to further segment the ``tail'' into ``subwords'' using words that already exist in the top portion of the vocabulary.

\paragraph{Step 1:} Use \texttt{THULAC} to process the corpus and acquire the vocabulary, rank it by frequency (likelihood), and choose a top portion of the vocabulary as the \emph{basic vocabulary}.

\paragraph{Step 2:} For the \emph{out-of-vocabulary} (OOV) words (i.e., the ``tail''), use the basic vocabulary to segment them into subwords. If there exists subwords outside the basic vocabulary, add them to the basic vocabulary. If an OOV word has different segmentation strategies, determine by calculating and choosing the largest likelihood product of subwords.

\paragraph{Step 3:} Sort the \emph{expanded} basic vocabulary again, and choose a top portion as the \emph{final vocabulary}.

After this, the coverage of the final vocabulary will be very close to 1 (see red curve in Figure \ref{fig:zipf} in Appendix \ref{appendix1}). This algorithm combines the advantages of CWS and frequency-based methods, and can generate a vocabulary with suitable size and high coverage. See Figure \ref{fig:vocab} in Appendix \ref{appendix1} for detailed illustrations.

\section{Diversity-aware Sampling}

\subsection{Controllable Diversity by Permutating the ``Head'' of Predicted Distribution}

As is widely applied for text generation, stochastic sampling methods (nucleus sampling, \citealp{holtzman2019curious} or top-\emph{k} sampling, ~\citealp{fan-etal-2018-hierarchical,holtzman-etal-2018-learning}) are able to overcome text degeneration issue and create better results than beam search algorithm. However, in poetry generation task we find that they \emph{still} generate boring and severely degenerated poetry (see Table \ref{tab:main_result} and Table \ref{fig:examples_baseline} in Appendix \ref{appendix1}).

Recent researches have revealed that low diversity issue are caused by the ``head'' part (high frequency) of the predicted distribution \citep{holtzman2019curious,basu2021mirostat,zhang2021improving}, which is not fully addressed by traditional stochastic sampling methods. \citet{zhang2021improving} propose that the permutation on the ``head'' can greatly improve the diversity of generated samples. This is in accordance with human instinct for poetry composing, that is, fluent sentences with too many high frequency words can be actually boring (i.e., the boredom trap, \citealp{basu2021mirostat}), while ambiguous sentences using surprising and low frequency words can actually be creative and poetic (e.g., poems by James Joyce or Marcel Proust).

%They propose that the ``head'' to permutate can be identified by interquartile range (IQR) calculation on the predicted distribution, which provides controllable diversity, as small value of IQR coefficient $\rho$ will expand the range of ``head'' to permutate hence increasing diversity, and vice versa (see Corollary 1, ~\citealp{zhang2021improving}).

%we first directly adopt the \emph{interquartile range inverse probability} (IQR-IP) sampling algorithm ~\citep{zhang2021improving} as a candidate baseline algorithm. Furthermore,

Inspired by this, we propose a simple sampling strategy to improve the novelty of poetry generation. To identify the high frequency ``head'', we propose to directly use nucleus sampling (NS) with a more strict parameter denoted by $q$ (stricter than the ``tail''). To avoid ``leakage of the tail'' (see Section 3.2, ~\citealp{zhang2021improving}), we adopt \emph{top1ctrl} filtering with parameter $n$ (Eq. 7, ~\citealp{zhang2021improving}) to prune the ``head'', in order to keep too-low-probability words outside the ``head''. The ``head'' denoted by $V^{head}$ is then jointly determined by $q$ and $n$ as follows.
\begin{equation}\label{eq:q_and_n}
\small
V^{head} = {NS}(q)\cap top1ctrl(n),
\end{equation}
\begin{equation}
\small
top1ctrl(n)=\big\{x\mid p(x)\ge\max{p(x)}/n\big\},
\end{equation}
where $p(x)$ denotes the auto-regressive predicted probability of word $x$. With ``head'' acquired, we propose to \emph{evenly redistribute} probability mass for $V^{head}$, in order to emphasize on the ``comparatively low frequency'' words in the ``head''. And for the ``tail'' (low frequency part) of the distribution, we directly adopt nucleus sampling with parameter $p$ ($p\ge q$) and truncate the ``tail'' like traditional methods. Stochastic sampling is finally conducted on the permutated distribution (with ``tail'' being cut off and ``head'' being randomized). The above method is referred to as \emph{nucleus sampling with randomized head} (NS-RH) algorithm. Clearly, the diversity gain of NS-RH algorithm is controlled by $q$ and $n$ which determine the boundary of $V^{head}$.
%We use these desirable features of these algorithms to control the diversity of the generated samples and alleviate text degeneration.

%We set $p$, $k$, and $n$ parameters of IQR-IP sampling algorithm to be fixed, and set IQR coefficient $\rho$ as the tunable diversity parameter, which is represented by the slider in the system interface in Figure \ref{tab:interface}.

\subsection{Semantic-similarity-based Rejection Sampling Algorithm}

Another challenge of poetry generation is the off-topic issue, since ``permutating the head'' essentially encourages to sampling on \emph{less probable} tokens and will practically contribute to the tangency from the topic. Also, in our scenario the input context (the poetry title or the keyword) is comparatively \emph{short and less informative}. Empirically, this is much more difficult than cases reported by \citet{holtzman2019curious}, as their context is considerably \emph{long and more informative} with a lot of human-provided details.
%By intuition, it will be beneficial to create a longer and more informative context on the basis of the short input title in order to alleviate the off-topic issue and lower the difficulty for later generation steps.

Inspired by the \emph{sentence-level rejection sampling algorithm} by \citet{kang-hashimoto-2020-improved} which is determined by perplexity, we propose the \emph{semantic-similarity-based rejection sampling algorithm} that involves the following 3 steps.

\paragraph{Step A:} Sample the first $M$ tokens for $N$ times and acquire $N$ \emph{hypothesis}.

\paragraph{Step B:} For each hypothesis, calculate the \emph{semantic similarity} between the hypothesis and the input context (the poetry title), then choose hypothesis with the highest semantic similarity as the \emph{accepted sample}.

\paragraph{Step C:} Feed the accepted sample into the model to generate all remaining tokens.

The key issue for semantic similarity in \textbf{Step B} is the calculation of \emph{sentence embedding}, as semantic similarity can be measured by the cosine distance between sentence embedding. We propose to adopt conclusions by ~\citet{zhang2021embedding} and use the average of hidden states for the last half number of Transformer layers (last 12 out of 24), together with \emph{standard normalization calibration} \citep{li-etal-2020-sentence} as the sentence embedding.

\section{Demonstration and Evaluation}
\begin{figure*}[hbt!]
\centering
\includegraphics[width=6in]{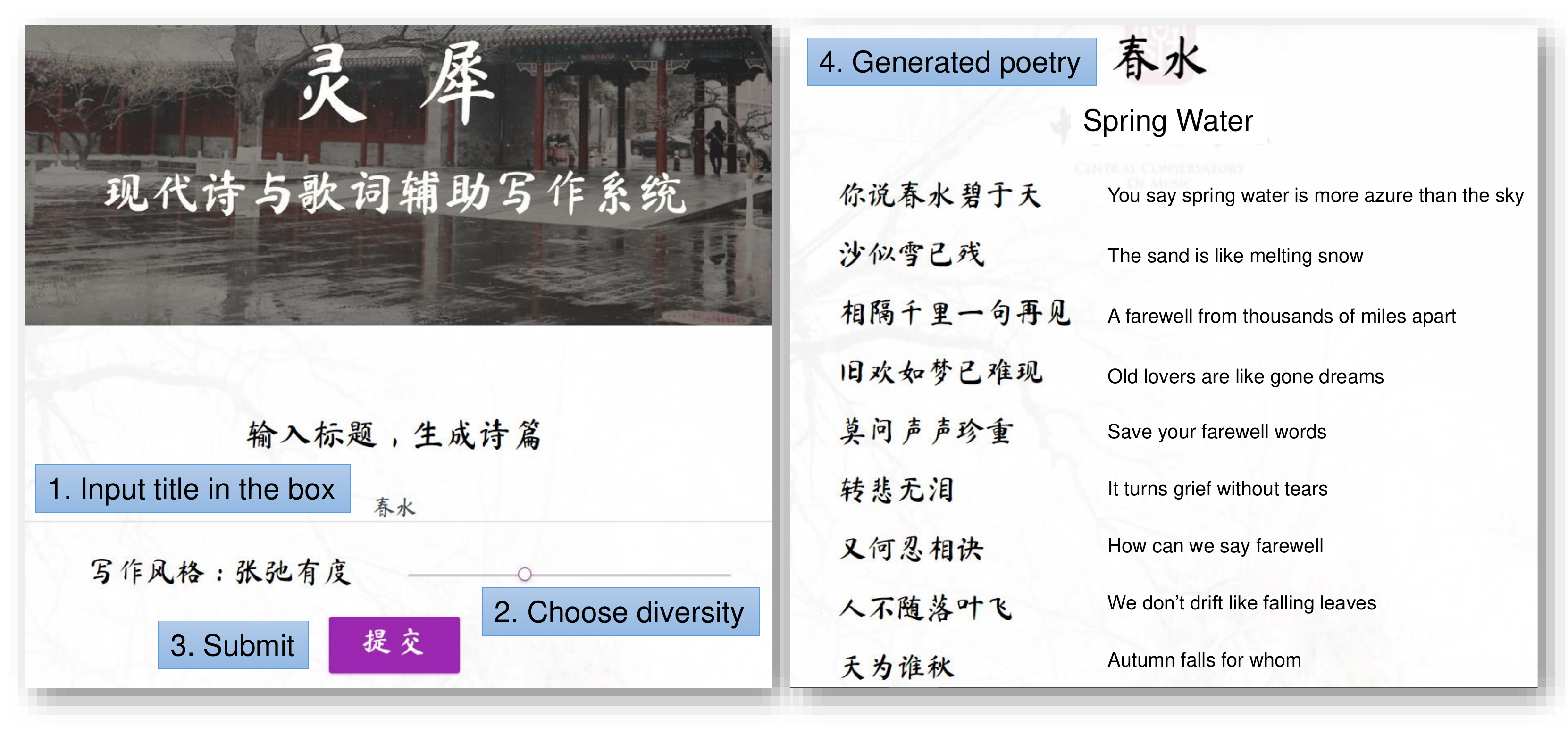}
\caption{System interface of \texttt{Lingxi} and poetry example (with translation).}\label{tab:interface}
\end{figure*}
\def\MyWidth{0.09\textwidth}
\begin{table*}[hbt!]
\tiny
\centering
\tabcolsep 1mm
\renewcommand{\arraystretch}{1.4}
\begin{center}
\begin{tabular}{
P{0.23\textwidth}P{\MyWidth}P{\MyWidth}P{\MyWidth}P{\MyWidth}P{\MyWidth}P{\MyWidth}P{\MyWidth}
}
\toprule
Method & PPL & Self-BLEU 4 & Self-BLEU 5 & Zipf Coef. & Rep. Entropy & Rhy. Entropy & Line Length \\
\midrule
Human & 16.75 & 0.45 & 0.32 & 0.90 & 3.68 & 1.50 & 7.42 \\
\midrule
NS, $p=0.70$ & 2.70 & 0.65 & 0.51 & \textbf{0.87} & 3.14 & \textbf{1.52} & 5.84  \\
NS, $p=0.90$ & 6.80 & 0.51 & 0.36 & 0.82 & 3.59 & 1.85 & 6.38 \\
Top-\emph{k}, $k=200$ & 7.31 & \textbf{0.50} & \textbf{0.35} & 0.81 & \textbf{3.61} & 1.86 & 6.66 \\
Pure sampling ($p=1.00$) & \textbf{18.83} & \textbf{0.40} & 0.25 & 0.80 & 3.87 & 2.03 & \textbf{7.05} \\
%Beam search (20 beams) & 25.14 & - & - & 1.24 & \textbf{4.52}  \\
\midrule
IQR-IP, $\rho=10.00$ & 6.60 & 0.55 & 0.38 & \textbf{0.81} & \textbf{3.70} & \textbf{1.85} & \textbf{7.24} \\
IQR-IP, $\rho=5.00$ & 9.54 & 0.51 & 0.34 & 0.76 & 3.86 & 1.88 & 7.90 \\
IQR-IP, $\rho=3.00$ & 12.13 & 0.48 & \textbf{0.32} & 0.73 & 3.96 & 1.91 & 8.29 \\
IQR-IP, $\rho=1.50$ & \textbf{18.14} & \textbf{0.43} & 0.27 & 0.70 & 4.08 & 1.92 & 9.33 \\
\midrule
%NS-RH, $p=0.80$, $q=0.10$ & 3.97 & 0.58 & 0.44 & \textbf{0.84} & 3.36 & \textbf{1.69} & 6.06 \\
NS-RH, $p=0.80$, $q=0.20$ & 4.00 & 0.58 & 0.43 & \textbf{0.84} & 3.37 & \textbf{1.70} & 6.09 \\
%NS-RH, $p=0.80$, $q=0.30$ & 4.53 & 0.56 & 0.41 & 0.85 & 3.43 & 1.72 & 6.15 \\
NS-RH, $p=0.80$, $q=0.40$ & 5.06 & 0.56 & 0.40 & 0.83 & 3.52 & 1.79 & 6.28 \\
%NS-RH, $p=0.80$, $q=0.50$ & 6.25 & 0.54 & 0.38 & 0.80 & \textbf{3.64} & 1.87 & 6.39 \\
NS-RH, $p=0.80$, $q=0.60$ & 8.82 & \textbf{0.50} & \textbf{0.34} & 0.78 & \textbf{3.80} & 1.95 & \textbf{6.69} \\
%NS-RH, $p=0.80$, $q=0.70$ & \textbf{14.69} & \textbf{0.44} & 0.28 & 0.73 & 3.99 & 2.00 & \textbf{7.26} \\
NS-RH, $p=0.80$, $q=0.80$ & \textbf{28.71} & 0.35 & 0.21 & 0.70 & 4.17 & 2.00 & 8.67 \\
\midrule
%NS-RH, $p=0.90$, $q=0.10$ & 6.82 & 0.51 & 0.36 & \textbf{0.82} & 3.59 & \textbf{1.85} & 6.38 \\
NS-RH, $p=0.90$, $q=0.20$ & 6.97 & 0.51 & 0.36 & \textbf{0.82} & 3.60 & \textbf{1.87} & 6.40 \\
%NS-RH, $p=0.90$, $q=0.30$ & 7.67 & 0.50 & 0.34 & 0.83 & \textbf{3.65} & 1.89 & 6.45 \\
NS-RH, $p=0.90$, $q=0.40$ & 8.80 & 0.49 & \textbf{0.33} & 0.81 & \textbf{3.72} & 1.93 & 6.51 \\
%NS-RH, $p=0.90$, $q=0.50$ & 10.47 & 0.47 & \textbf{0.31} & 0.78 & 3.82 & 1.99 & 6.74 \\
NS-RH, $p=0.90$, $q=0.60$ & \textbf{14.45} & \textbf{0.44} & 0.27 & 0.75 & 3.95 & 2.02 & \textbf{7.08} \\
%NS-RH, $p=0.90$, $q=0.70$ & 22.92 & 0.38 & 0.23 & 0.71 & 4.10 & 2.04 & \textbf{7.72} \\
NS-RH, $p=0.90$, $q=0.80$ & 41.54 & 0.30 & 0.17 & 0.68 & 4.24 & 2.01 & 9.25 \\
\bottomrule
\end{tabular}
\end{center}
\caption{Statistical evaluation for selected decoding parameters (\emph{metric closer to human text is better and in bold}). Lower self-BLEU score, lower Zipf coefficient and higher repetition entropy indicates \emph{higher diversity}.}\label{tab:main_result}
\end{table*}
\subsection{System Interface}
%Public domain name address for the system will be released later.
\footnotetext{\textsuperscript{1}Video demonstration is available at \url{https://youtu.be/rVp7jsCyu08}.}
%\texttt{Lingxi} is available at \url{http://166.111.5.187:16060/lingxi}\textsuperscript{1}.
The system interface and poetry example are shown in Figure \ref{tab:interface}. Its usage is very simple. First, type the input poetry title in the box, and then choose the diversity parameter with the slider. Sliding to the right will create more diversified samples, and vice versa. We set 4 levels of diversity parameters as is presented in Table \ref{tab:network}, \ref{fig:examples_text} and \ref{fig:examples_rh} in Appendix \ref{appendix1}.

\subsection{Statistical Evaluation and Diversity-aware Samples}
Statistical evaluation results are presented in Table \ref{tab:main_result}. We follow the evaluation paradigm by ~\citet{holtzman2019curious}, which calculates the corresponding metrics on the \emph{generated} samples by different sampling methods to compare against metrics of human text (closer metric to human is better). Perplexity (PPL) reflects general fluency of the generated sample (lower score indicates higher fluency but more boredom). Self-BLEU score \citep{holtzman2019curious,10.1145/3209978.3210080} reflects diversity among different samples (lower score indicates higher diversity). Zipf coefficient \citep{Zipf49,newman2005power} reflects the word frequency distribution feature (lower score indicates flatter word frequency distribution and higher diversity). Repetition entropy is calculated by $\mathbb{E}\{-\log{p(x)}\}$, where $p(x)$ is the word frequency distribution in the generated sample (higher score indicates less repetition and higher diversity). Rhyming entropy is calculated by $\mathbb{E}\{-\log{p_{rhyme}(x)}\}$, where $p_{rhyme}(x)$ is the rhyme frequency distribution in the generated sample (higher score indicates higher diversity but less rhymed). And line length is the average length of Chinese words on each poetry line in the generated sample (high score indicates more lengthy and informative poetry). We generate $5,000$ samples for each sampling method to calculate these metrics.

For PPL evaluation in Table \ref{tab:main_result}, results show that samples generated by traditional stochastic sampling method are \emph{severely degenerated}, as their perplexity (NS, $p=0.70$ or $0.90$ and top-\emph{k}, $k=200$) are far less than human metric. Clearly, they favor high frequency (high probability) words that result in lower PPL, which is extremely harmful for poetry generation, since a ``too fluent'' poetry with lower PPL can be easily regarded as poorly written, while semantically ambiguous sentences with higher PPL can be possibly regarded as poetic. \emph{Only} when $p=1.00$ can they achieve maximum PPL near the human metric, which is done by letting in all ``tails'' of the distribution. This is already proven to be a bad choice ~\citep{holtzman2019curious} because ``tails'' contain low probability words that might be unreasonable and corrupt the quality of the poetry. It also reveals the upper bound of diversity using traditional methods. For comparison, by using NS-RH algorithm that ``permutate the head'', the PPL significantly increases to human level without letting in more ``tails''. It also shows that the diversity gain is controllable by the diversity parameter ($q$ of NS-RH algorithm or $\rho$ of IQR-IP sampling algorithm by ~\citealp{zhang2021improving}), which controls the intensity of permutation.

Note a very interesting behavior of NS-RH algorithm, where NS-RH with $p=0.90$, $q=0.80$ achieves 41.54 of PPL and still generates satisfactory samples (see Table \ref{fig:examples_rh} in Appendix \ref{appendix1}). This means that even if almost top $80\%$ of the vocabulary is randomized, it \emph{still} generated fluent poetry but with higher novelty than traditional sampling methods. Clearly, this is because poetry requires higher diversity and novelty (favoring less probable words) than plain text generation task. It also shows that setting $p=0.80$, $q=0.80$ (filtering in top $80\%$ of the vocabulary and almost completely randomizing them) results in 28.71 of PPL, which is even closer to human PPL. This indicates that the filtered vocabulary using nucleus sampling can be \emph{almost completely randomized} but actually results in better metrics. This suggests that in cases that require high novelty, the value of likelihood on a ``flat'' distribution predicted by the language model can be \emph{unreliable} for creative generation, while completely randomizing the filtered vocabulary and ignoring the originally predicted likelihood during sampling can actually achieve better results with higher novelty. Note that this conclusion only applies for nucleus sampling since it adaptively works between ``flat distribution'' and ``peaked distribution'' (see results by ~\citealp{holtzman2019curious}). By inference, such method can be very suitable for other artistic generation tasks like music or drawings that requires high novelty.

Analyses for other metrics are very similar to PPL. For self-BLEU score, our algorithm achieves human metric without letting in more ``tails'' like traditional methods. For Zipf coefficient, our method can achieve much lower metric that indicates flatter and more diverse distribution of words, which can't be achieved by traditional methods. For repetition, our methods are less repetitive than traditional methods (with higher Rep. entropy). For rhyming, it shows our method will sacrifice rhyming (with higher Rhy. entropy than human metric) for diversity gain. And for line length, our algorithm can achieve similarly formatted poetry to human text, while traditional methods can't.

We present samples by traditional methods and our methods in Table \ref{fig:examples_baseline}, \ref{fig:examples_text} and \ref{fig:examples_rh} in Appendix \ref{appendix1}. It shows that \emph{less diversified} parameters ($q=0.20$ or $\rho=10.00$) will create plain sentences which resemble more to baseline algorithms, while \emph{highly diversified} ($q=0.80$ or $\rho=1.50$) will create diversified sentences with notably higher novelty, achieving diversity-aware sampling.
%Such tunable diversity can help to provide inspirations for the users to further modify and create their own desirable poetry on the basis of the generated samples.

%Table \ref{fig:examples_text} also shows that baseline algorithms (top-\emph{p}=$0.70$ or $0.90$, top-\emph{k}=$200$) tend to generate degenerated samples with low diversity. Obviously this is because these methods encourage to sample/decode on high-probability (high-likelihood and high-frequency) words, while these words can be practically tedious and repetitive. Continuingly sampling in this way will result in extreme cases such as repetition loops (see Figure 1, ~\citealp{zhang2021improving}). These behaviors are very different from human text behavior which does \emph{not} always choose high-probability candidates (see results by ~\citealp{holtzman2019curious}). Thus they can be addressed by encouraging sampling on less probable tokens.

\subsection{Ablation of Rejection Sampling}
\begin{table}[hbt!]
\small
\centering
\tabcolsep 1mm
\renewcommand{\arraystretch}{1.4}
\begin{center}
\begin{tabular}{
p{0.14\textwidth}
P{0.12\textwidth}P{0.12\textwidth}
}
\toprule
\multirow{2}{*}[-0.4em]{Diversity}&\multicolumn{2}{c}{BLEU $\uparrow$ ($\times 0.01$)}\\
\cmidrule{2-3}
& w/ RJ & w/o RJ \\
\midrule
$\rho=10.00$ & 0.67 & 0.50 \\
$\rho=5.00$ & 0.67 & 0.45 \\
$\rho=3.00$ & 0.65 & 0.39 \\
$\rho=1.50$ & 0.62 & 0.34 \\
\midrule
$q=0.20$ & 0.97 & 0.56 \\
$q=0.40$ & 0.70 & 0.41 \\
$q=0.60$ & 0.53 & 0.31 \\
$q=0.80$ & 0.37 & 0.23 \\
\bottomrule
\end{tabular}
\end{center}
\caption{Ablation of rejection sampling algorithm. It shows that with rejection sampling, generated poetry will always have higher BLEU score against the input poetry title, which helps the generated poetry to be on topic with the poetry title.}\label{tab:title_bleu}
\end{table}

\begin{figure}[hbt!]
\centering
\includegraphics[width=3in]{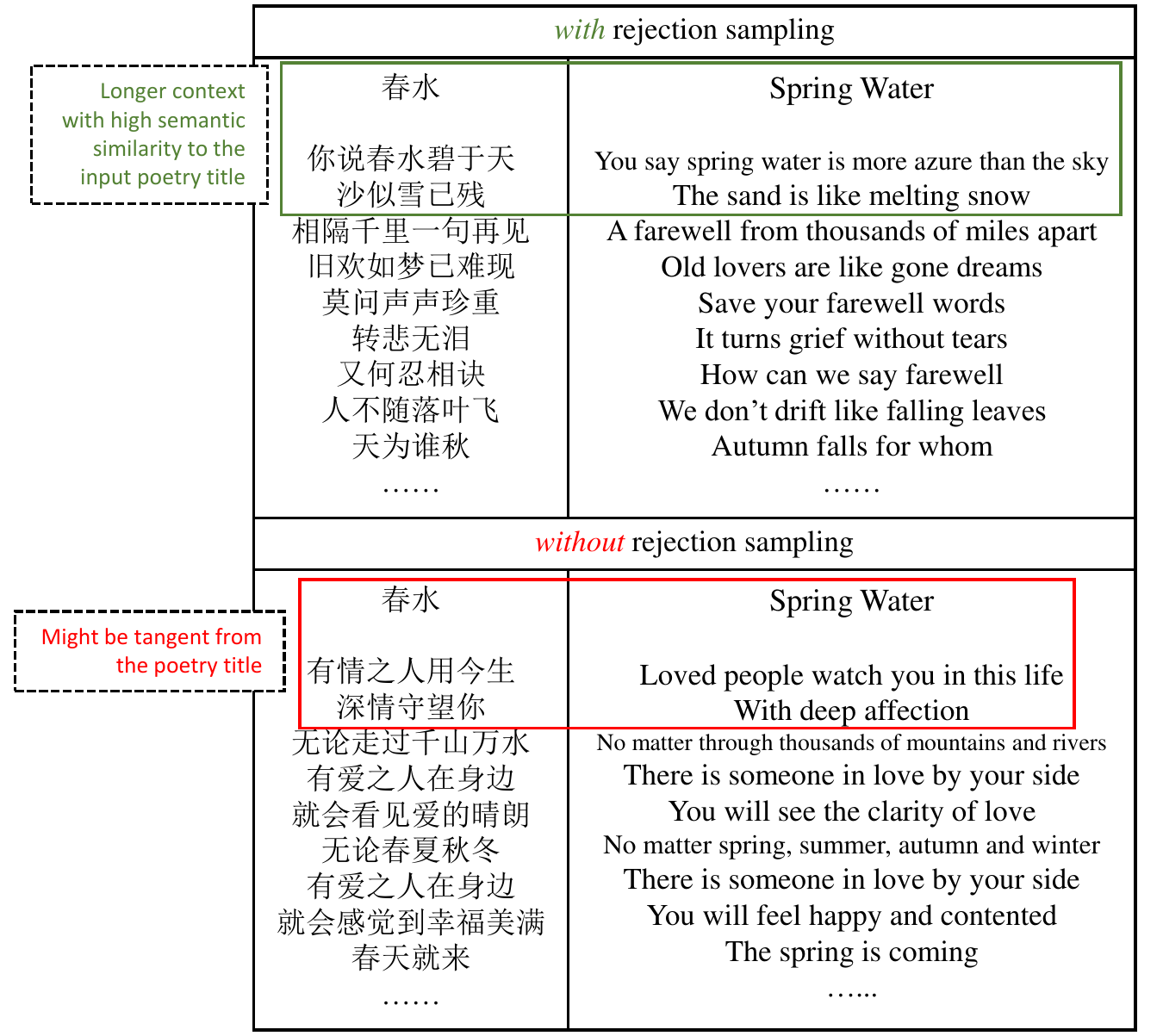}
\caption{Illustration of rejection sampling. By using rejection sampling, the context will be longer and more informative with high semantic similarity to the input poetry title.}\label{fig:examples_ablation}
\end{figure}
\begin{table*}[hbt]
\small
\centering
\tabcolsep 1mm
\renewcommand{\arraystretch}{1.4}
\begin{center}
\begin{tabular}{
p{0.35\textwidth}
P{0.12\textwidth}P{0.12\textwidth}P{0.12\textwidth}P{0.12\textwidth}
}
\toprule
System & Fluency & Novelty & Coherence & Overall \\
\midrule
Youling \citep{zhang-etal-2020-youling} & 4.37 & 4.16 & 4.19 & 4.24\\
XiaoIce \citep{cheng2018image} & \textbf{4.46} & 4.12 & 4.12 & 4.23\\
%\texttt{Lingxi} (\emph{ours}, $\rho=3.00$) & 4.22 & \textbf{4.43} & \textbf{4.16} & \textbf{4.27}\\
\texttt{Lingxi} (\emph{ours}, $q=0.60$) & 4.43 & \textbf{4.31} & \textbf{4.23} & \textbf{4.32}\\
\bottomrule
\end{tabular}
\end{center}
\caption{Human evaluation results for publicly available Chinese modern poetry generation systems. It shows that our system achieves similar fluency but higher novelty and coherence compared with other systems.}\label{tab:human}
\end{table*}
%Since the input poetry title is comparatively shorter and less informative than plain generates tasks such as cases reported by \citet{holtzman2019curious}, empirically it will more easily result in off-topic in later generation steps.
We present an ablation study for the rejection sampling algorithm in Table \ref{tab:title_bleu}. To explore different settings, we first adopt IQR-IP sampling algorithm and fix $\rho=5.00$ in step \textbf{A} to acquire each hypothesis, i.e., with \emph{fixed} diversity for hypothesis, then use tunable $\rho$ in step \textbf{C} for observation. For comparison, for NS-RH algorithm we directly use identical parameters for step \textbf{A} and \textbf{C} with tunable $q$, i.e., using \emph{tunable} diversity for hypothesis. The evaluated metric (BLEU score) is calculated by treating the sharing input poetry title as the reference and treating each generated poetry as the hypothesis. Results are very clear that with rejection sampling, the achieved BLEU score is always higher. For IQR-IP sampling algorithm, since we use fixed and \emph{lightly diversified} parameters in step \textbf{A}, the score with rejection sampling does not vary dramatically with different diversity parameters. And even in the setting for NS-RH algorithm that uses tunable diversity parameters for step \textbf{A}, the score with rejection sampling is still higher than cases without it. This means that the proposed rejection sampling algorithm can help the generated poetry to be on topic in different diversity levels and different settings. We present a pair of samples in Figure \ref{fig:examples_ablation} to illustrate the impact of the proposed algorithm.
%See Figure \ref{fig:examples_ablation} in Appendix \ref{appendix1} for example.

%Clearly, rejection sampling (on the left side) helps to generate longer context on the basis of the short poetry title (inside the green box) with high semantic similarity to the title, while without rejection sampling the first few tokens (inside the red box) might be less relevant to the poetry title, which might possibly lead to tangency for the whole passage.
%Results in Table \ref{tab:human} also show that poetry from our system has better consistency. These results indicate that the proposed rejection sampling can create more informative context that lowers the risk of being off-topic.

\subsection{Human Evaluation}
\footnotetext{\textsuperscript{2}\url{https://yl.fuxi.netease.com/}}
\footnotetext{\textsuperscript{3}\url{https://poem.msxiaobing.com/}. XiaoIce requires image input, so we choose an art drawing from \url{https://artexpress.artron.net/wap/works/detail?works_id=245462} composed with the same title (Spring Water) as image input, as well as the poetry title itself as keyword input. We choose the longest version of XiaoIce's output for comparison.}

We provide human evaluation comparing with other publicly available Chinese modern poetry generation system (Youling\textsuperscript{2} by \citealp{zhang-etal-2020-youling} and XiaoIce\textsuperscript{3} by \citealp{cheng2018image}). Since these systems do not provide open-source model and data, and have different functionalities (e.g., attribute control function of Youling, and image input function of XiaoIce), it is difficult to directly compare with them. We consider the primary feature of our system that addresses novelty and coherence, so we design human evaluation regarding these features for comparison. We generate 50 pieces of poetry from each system (\texttt{Lingxi} with $q=0.60$) under the same input poetry title (``spring water'' in Chinese), and collect human annotations on \emph{fluency} (focusing on grammaticality and linguistic clarity), \emph{novelty} (focusing on the extent of being poetic and creative) and \emph{coherence} (focusing on consistency to the title and the poetry context) on a 1 to 5 scale (larger better). Annotators include: 1) Chinese graduate students and advisors of our research team who are familiar with neural generation problem, and 2) professional musicians and song composers who are familiar with Chinese lyric composing. They are presented with randomly chosen samples, while being unaware of the actual poetry system during annotation, and are required to rate these samples by their instinct without too much reconsideration. We receive 103 annotations from 9 different annotators for each system. Results are shown in Table \ref{tab:human}. It shows that \texttt{Lingxi} generates poetry with considerable novelty and coherence gain while maintaining satisfactory fluency comparing with other systems. Clearly, these features are unavailable in baseline systems and unable to achieve by traditional sampling methods.

%Table \ref{tab:others} shows that other systems might generate comparatively less diversified poetry with occasional repetition (like Youling) or less surprising sentences (like XiaoIce), while poetry of our system has lower repetition and notably \emph{different literal style} with satisfactory fluency, due to IQR-IP sampling algorithm that permutates the distribution and avoids always sampling on high-probability candidates.

\section{Conclusion}

In this work we present \texttt{Lingxi}, a diversity-aware Chinese modern poetry generation system. We propose nucleus sampling with randomized head (NS-RH) algorithm that achieves controllable diversity. We also propose a semantic-similarity-based rejection sampling algorithm to alleviate the off-topic issue. Our results indicate that randomizing the high frequency part of the distribution can significantly increase the novelty of generated poetry. For artistic generation that requires greater novelty, our method can achieve better results than directly sampling on the original distribution.
%For future works, combining multimodal feature (e.g., image, speech or music) would be an interesting research option.

%\section*{Acknowledgments}
%
%The acknowledgments should go immediately before the references. Do not number the acknowledgments section.
%\textbf{Do not include this section when submitting your paper for review.}

\bibliographystyle{acl_natbib}
\bibliography{poetry2020}

\clearpage
\appendix

\section{Model Configuration, Training Details and Generated Examples}\label{appendix1}
\begin{figure}[hbt!]
\centering
\includegraphics[width=3in]{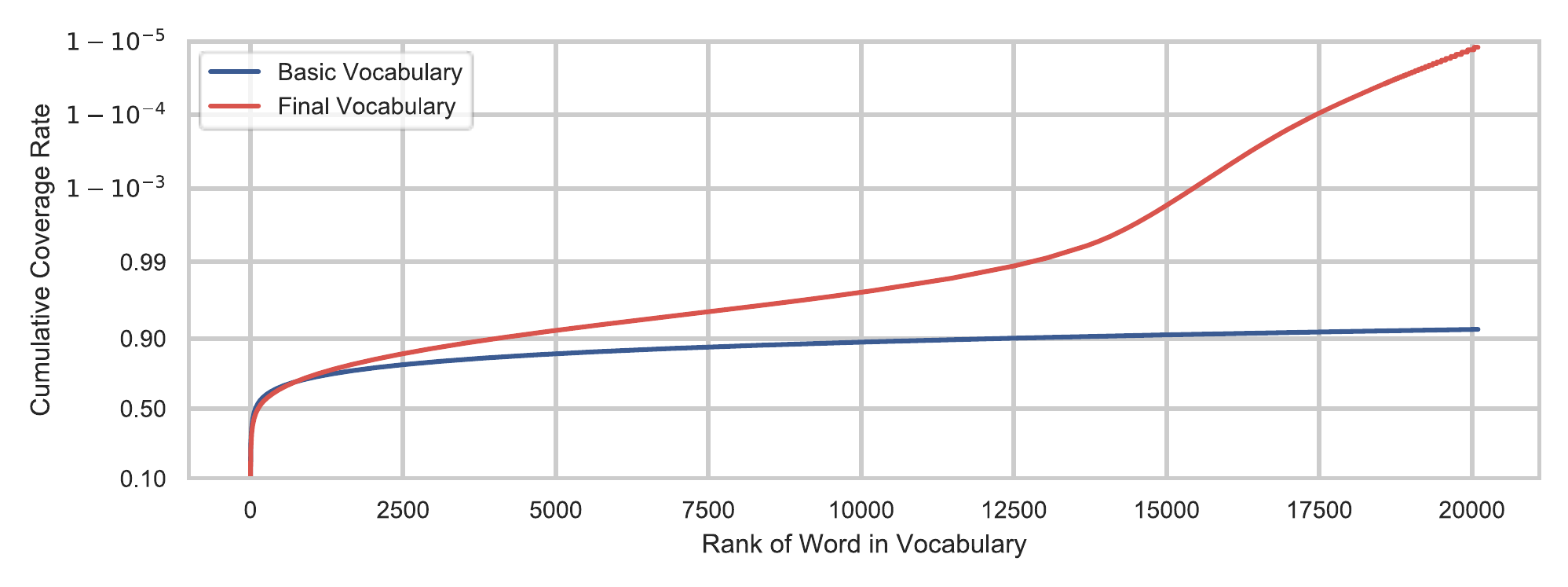}
\caption{Comparison of vocabulary coverage. By using the proposed segmentation algorithm, vocabulary coverage is increased while maintaining a suitable vocabulary size.}\label{fig:zipf}
\end{figure}
\begin{figure*}[hbt!]
\centering
\includegraphics[width=6.3in]{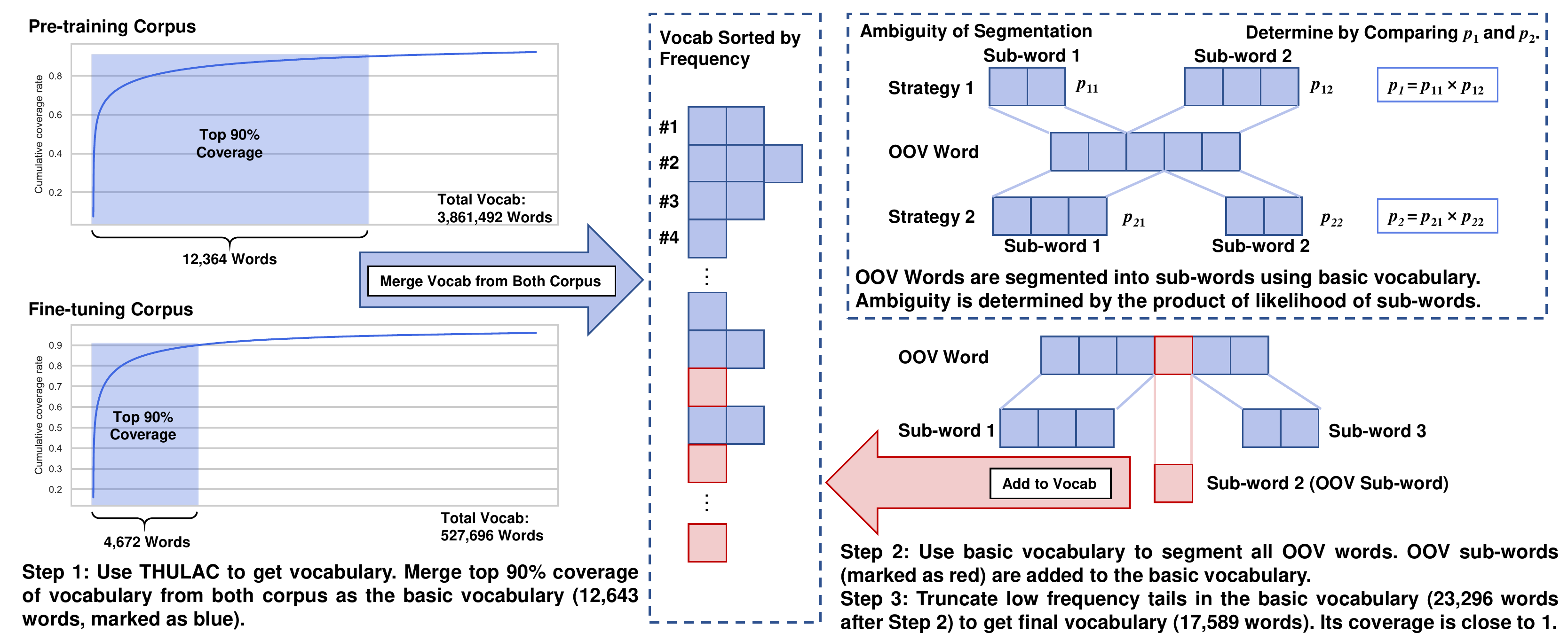}
\caption{Illustration of the corpus preprocessing.}\label{fig:vocab}
\end{figure*}

\begin{table*}[htb!]
\centering
%\vspace*{-1 cm}
\tabcolsep 1mm
\begin{center}
\begin{tabular}{
p{0.50\textwidth}
p{0.40\textwidth}
}
\toprule
Parameters & Value \\
\midrule
number of Transformer layers & 24 \\
number of Transformer attention heads & 16 \\
%number of Transformer heads & 16 \\
embedding size & 1,024 \\
vocabulary size & 17,589 \\
maximum context length & 128 \\
number of network parameters & 330 million \\
pre-training epochs & 20 \\
fine-tuning epochs & 10 \\
batch size per GPU& 32 \\
number of training GPUs & 8 NVIDIA\textsuperscript{\textregistered} GeForce\textsuperscript{\textregistered} RTX 2080 Ti\\
pre-training learning rate & $2\times 10^{-4}$\\
fine-tuning learning rate & $\{1\times 10^{-5}, 2\times 10^{-5}\}$\\
learning rate decay & linear decay \\
warm up steps & 1\% of total steps\\
optimizer & Adam optimizer \citep*{kingma2014adam} \\
weight decay & 0.01 \\
PPL on valid set after pre-training & 17.58\\
best fine-tuning epoch & epoch 4 with $2\times 10^{-5}$ learning rate\\
best PPL on fine-tuning valid set & 16.75\\
rejection sampling parameters & $N=16$, $M=20$ \\
Sampling parameters for IQR-IP sampling algorithm & $p=0.90$, $k=200$, $n=100$, tunable $\rho$\\
Sampling parameters for NS-RH algorithm & $p=0.90$, $n=100$, tunable $q$ \\
%IQR coefficient levels in \textbf{Step C} & $\rho\in\{1.50, 3.00, 5.00, 10.00\}$\\
\bottomrule
\end{tabular}
\caption{Model configuration, training details and sampling parameters}\label{tab:network}
\end{center}
\end{table*}
Detailed configurations and parameters of our model are listed in Table \ref{tab:network}. The preprocessing of the corpus is illustrated in Figure \ref{fig:vocab}. Since we have two different training corpora, in order to exploit word frequency feature from both corpora, in Section \ref{preprocessing} for \textbf{Step 1} we choose the top 90\% of the vocabulary from: a) both corpora, and b) the fine-tuning corpus only, then merge them as the basic vocabulary. This emphasizes on the fine-tuning corpus in order to help the final generation task. The size of basic vocabulary merged in \textbf{Step 1} is 12,643. After \textbf{Step 2}, the vocabulary size is expanded to 23,296. In \textbf{Step 3}, we drop words with frequency lower than: a) 100 on both corpora, and b) 10 on the fine-tuning corpus, and acquire the final vocabulary with 17,589 words. By observation, the dropped words in \textbf{Step 3} are all extremely rare single-length Chinese words with very low frequency.

Figure \ref{fig:zipf} illustrates the cumulative coverage of our vocabulary, which is in accordance with the well-known Zipf's law \citep{Zipf49,newman2005power}, as a small portion of high frequency words (e.g., first 20,000 words in Figure \ref{fig:zipf}, ranked by word frequency) will take a large portion of coverage in the corpus (e.g., about \%90 coverage in Figure \ref{fig:zipf}), while the long ``tail'' with low frequency words will take a small portion of coverage (e.g., total 3.6 million minus first 20,000 words in the pre-training corpus takes the remaining \%10 coverage in Figure \ref{fig:zipf}).

For special tokens, the [\texttt{TITLE}] token is added directly after the title of each poetry passage in the fine-tuning corpus to capture the title feature. The [\texttt{START-OF-PASSAGE}] token is added before the starting token of the first poetry line. We create a replica for each poetry passage excluding the title and [\texttt{TITLE}] token, and mix them with the original corpus as data augmentation. The [\texttt{NEWLINE}] token is added at the end of each poetry line in replace of newline character, and the [\texttt{END-OF-PASSAGE}] token is added at the end of each poetry passage. English words and letters are assigned [\texttt{UNK-EW}] tokens. Other unknown words and sub-words are assigned [\texttt{UNK}] tokens. For poetry passages longer than maximum context length, we create training samples using a sliding window with stride being half of maximum context length. We split train/valid/test sets using the common ratio of 85\%/7\%/8\% (token ratio for the pre-training corpus, passage ratio for the fine-tuning corpus).

The model is an auto-regressive Transformer decoder, using language model loss as training loss. It achieves monotonic convergence of perplexity (PPL) on the valid set of the pre-training corpus at the end of the pre-training steps. We choose the best fine-tuning epoch of model with the lowest PPL on the valid set of fine-tuning corpus as the final model. The achieved metrics are reported in Table \ref{tab:network}. The training program is developed using the library released by ~\citet{Wolf2019HuggingFacesTS}. The sliding window strategy for PPL calculation also follows methods described in their official documents.

\begin{table*}[hbt!]
\centering
\includegraphics[width=6in]{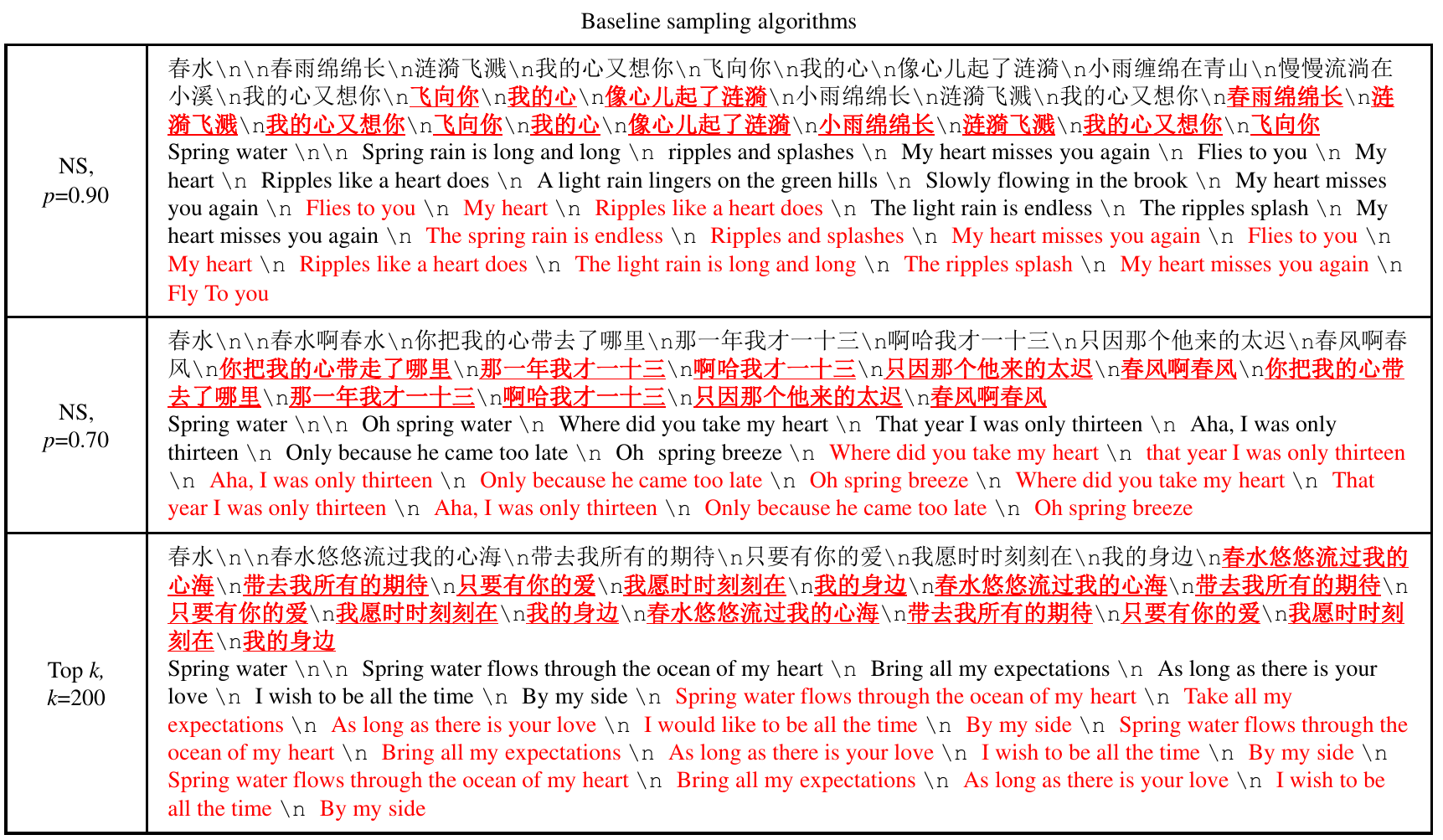}
\caption{Samples decoded by traditional stochastic sampling methods. Repetitions are marked with red underline in bold. It shows that even with stochastic sampling algorithms, the model still tends to generate repetitive and boring sentences with low novelty (\emph{degenerated}).}\label{fig:examples_baseline}
\end{table*}

\begin{table*}[hbt!]
\centering
\includegraphics[width=6in]{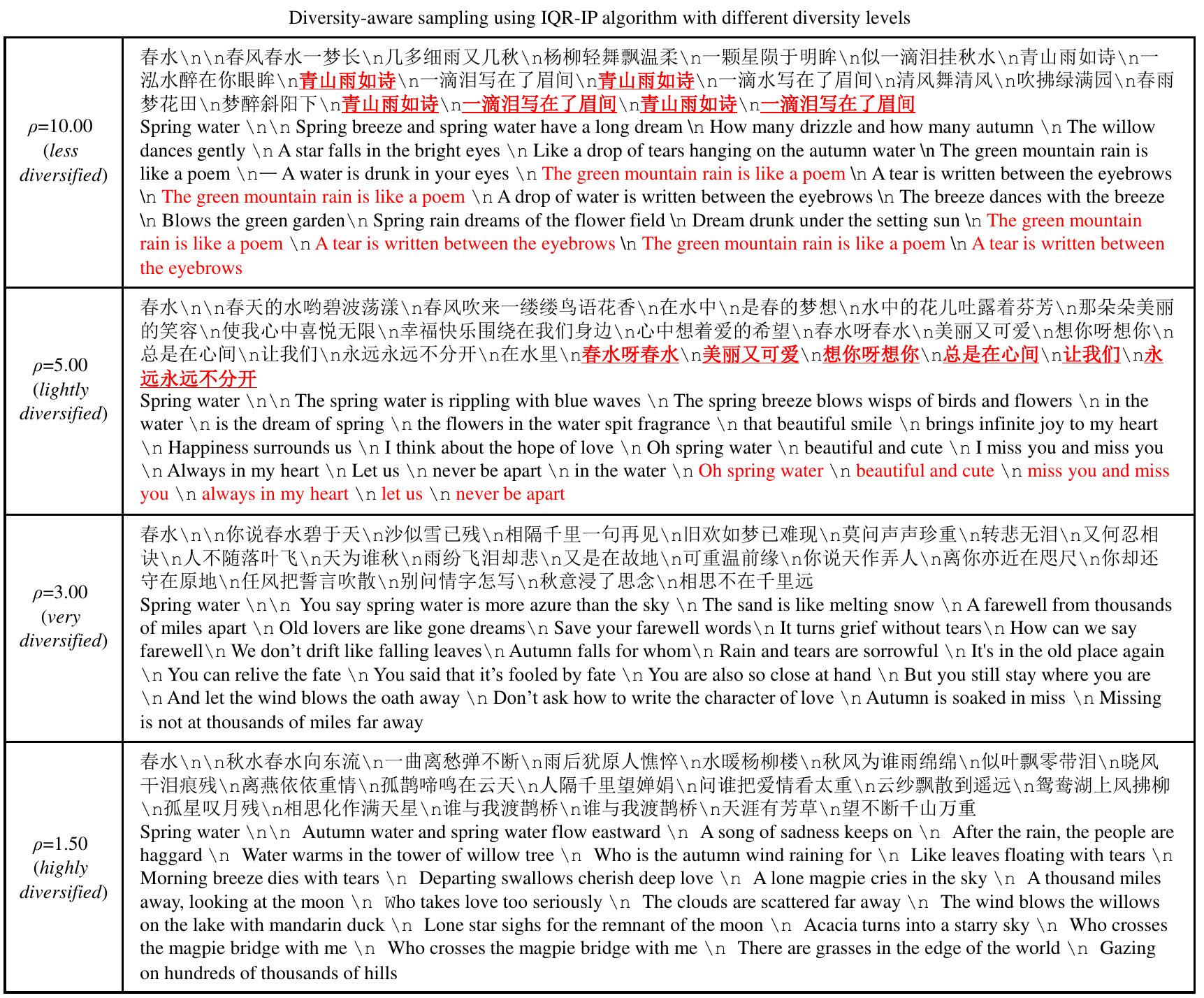}
\caption{Diversity-aware samples using IQR-IP algorithm. Repetitions are marked with red underline in bold. It shows that ``permutating the head'' can achieve controllable diversity by using different diversity parameter, alleviating the degeneration issue and maintaining satisfactory fluency.}\label{fig:examples_text}
\end{table*}

\begin{table*}[hbt!]
\centering
\includegraphics[width=6in]{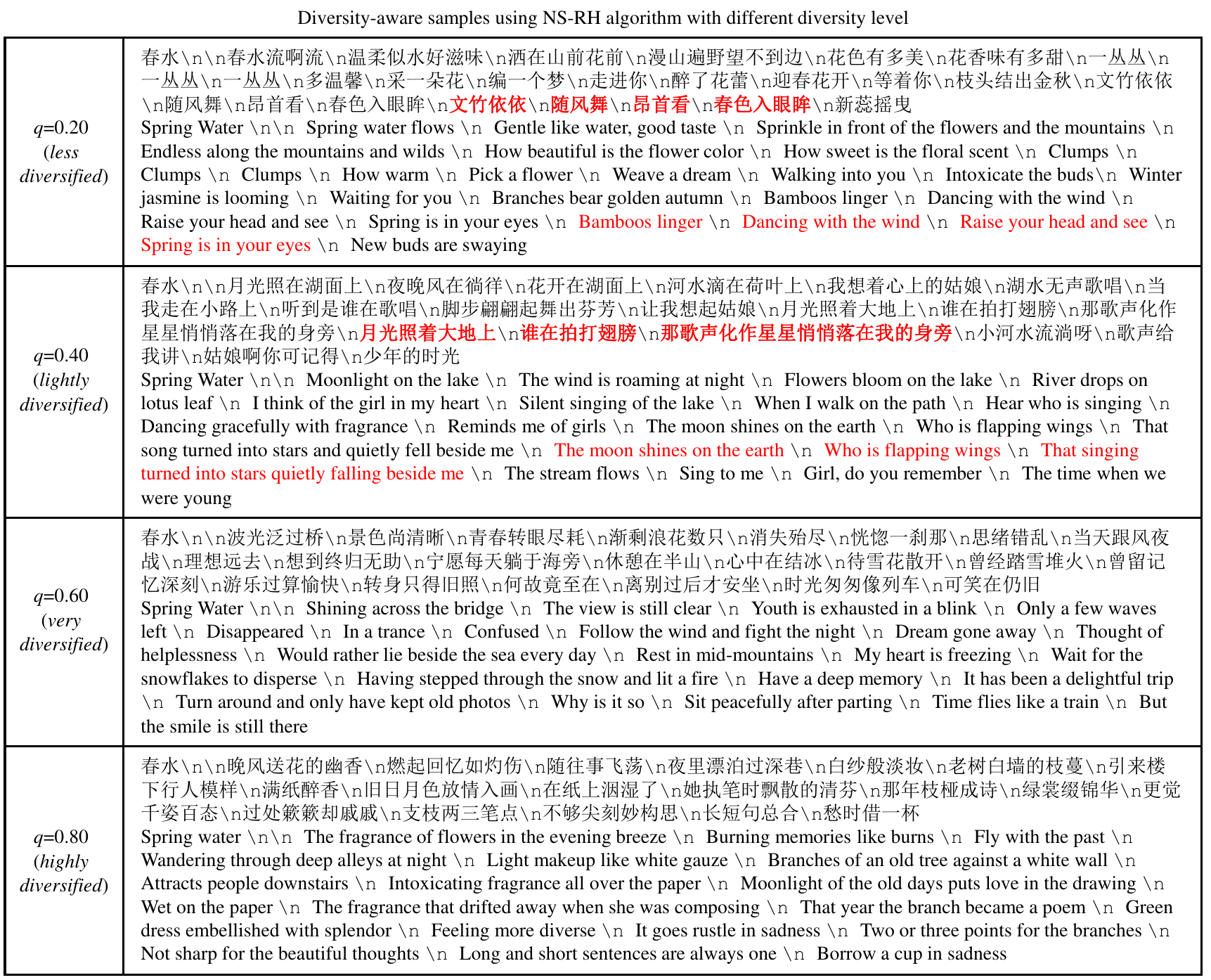}
\caption{Diversity-aware samples using NS-RH algorithm. Repetitions are marked with red underline in bold. It shows that even with $q=0.80$ which means almost top $80\%$ of the vocabulary is randomized, our system still generated fluent samples, but with notably higher novelty and creativity compared with traditional degenerated samples.}\label{fig:examples_rh}
\end{table*}

%\begin{table*}[hbt!]
%\centering
%\includegraphics[width=4.5in]{examples_youling.pdf}
%\caption{Examples for comparison with other publicly available Chinese modern poetry generation systems. Poetry from our system is less repetitive, less wordy and more diversified compared with other systems.}\label{tab:others}
%\end{table*}

\end{document}